\documentclass{article}

\usepackage[final]{nips_2016}
\usepackage[nottoc,notlof,notlot]{tocbibind}
\usepackage{hyperref}
\usepackage{graphicx}
\usepackage{amssymb}
\usepackage{textcomp}
\usepackage{amsmath}
\usepackage[utf8]{inputenc} %
\usepackage[T1]{fontenc}    %
\usepackage{hyperref}       %
\usepackage{url}            %
\usepackage{booktabs}       %
\usepackage{amsfonts}       %
\usepackage{comment}
\usepackage{nicefrac}       %
\usepackage{microtype}      %
\usepackage{caption}
\usepackage{subcaption}

\newenvironment{definition}[1][Definition]{\begin{trivlist}
\item[\hskip \labelsep {\bfseries #1}]}{\end{trivlist}}

\newcommand{\qed}{\nobreak \ifvmode \relax \else
      \ifdim\lastskip<1.5em \hskip-\lastskip
      \hskip1.5em plus0em minus0.5em \fi \nobreak
      \vrule height0.75em width0.5em depth0.25em\fi}

\title{Machine Translation: A Literature Review}

\author{
  Ankush Garg, Mayank Agarwal\\
  Department of Computer Science\\
  University of Massachusetts Amherst \\
  \texttt\{agarg,mayankagarwa\}@cs.umass.edu 
}

\begin{document}

\maketitle

\begin{abstract}

Machine translation (MT) plays an important role in benefiting linguists, sociologists, computer scientists, etc. by processing natural language to translate it into some other natural language. And this demand has grown exponentially over past couple of years, considering the enormous exchange of information between different regions with different regional languages. Machine Translation poses numerous challenges, some of which are: a) Not all words in one language has equivalent word in another language b) Two given languages may have completely different structures c) Words can have more than one meaning. Owing to these challenges, along with many others, MT has been active area of research for more than five decades. Numerous methods have been proposed in the past which either aim at improving the quality of the translations generated by them, or study the robustness of these systems by measuring their performance on many different languages. In this literature review, we discuss statistical approaches (in particular word-based and phrase-based) and neural approaches which have gained widespread prominence owing to their state-of-the-art results across multiple major languages.

\end{abstract}

\section{Introduction}

Machine Translation is a sub-field of computational linguistics that aims to automatically translate text from one language to another using a computing device. To the best of our knowledge, Petr Petrovich Troyanskii was the first person to formally introduce machine translation \cite{hutchins2000petr}. In 1939, Petr approached the Academy of Sciences with proposals for mechanical translation, but barring preliminary discussions these proposals were never worked upon. Thereafter, in 1949, Warren Weaver \cite{weaver1955translation} proposed using computers to solve the task of machine translation. Since then, machine translation has been studied extensively under different paradigms over the years. Earlier research focused on rule-based systems, which gave way to example-based systems in the 1980s. Statistical machine translation gained prominence starting late 1980s, and different word-based and phrase-based techniques requiring little to no linguistic information were introduced. With the advent of deep neural networks in 2012, application of these neural networks in machine translation systems became a major area of research. Recently, researchers announced achieving human parity on automatic chinese to english news translation \cite{hassan2018achieving} using neural machine translation. While early machine translation systems were primarily used to translate scientific and technical documents, contemporary applications are varied. These include various online translation systems for exchange of bilingual information, teaching systems, and many others.

In this literature review, we survey two major sub-fields of machine translation: statistical machine translation, and neural machine translation. The rest of the review is structured as follows: Section 2 briefly discusses the early work in machine translation. Section 3 reviews Statistical machine translation, focusing on word-based and phrase-based translation techniques. Section 4 elaborates on neural machine translation techniques where we also discuss different attention mechanisms and architectures with special purposes. Section 5 briefly describes the current research in the field. Finally, we conclude the report with Section 6.

\section{Early Work}

In the 1970's, Rule-based Machine Translation (RBMT) was the primary focus of research.  Such systems fall into one of the following three categories: Direct systems (these map input sentence directly to the output sentence), Transfer RBMT systems (these use morphological and syntactic analysis to translate sentences), and Interlingual RBMT systems (these transformed the input sentence to an abstract representation and mapped this abstract representation to the final output). One such work in Interlingual RBMT system is by Carbonell \textit{et al.} in 1978 \cite{carbonell1978knowledge}. The proposed approach translates text by: 1) Converting the source text to a language-free conceptual representation, 2) Augmenting this representation with information that was implicit in the source text, and 3) Converting this augmented representation to the target language. The authors argue that translation requires detailed understanding of the source text which semantic rules are inadequate to capture and therefore need to be augmented with detailed domain knowledge as well.

Rule-based MT is complicated for certain languages (ex: English-Japan) owing to different structures of the languages. In 1984, Nagao \cite{nagao1984framework} proposed a translation system that works by analogy principle. Titled "machine translation by example-guided inference", the system relies on a big dataset of example sentences and their translations to learn the correspondence between English-Japanese words and also the structure of the language. The authors describe different approaches to build such a system and also discuss ways to curate the data required for such a system. This paper, to the best of our knowledge, is the first paper to introduce example-based learning and paved way for further research in building machine translation systems that do not rely on manually curated rules and exceptions.

\section{Statistical Machine Translation}

Statistical Machine Translation (SMT), as introduced by Brown \textit{et al.} \cite{brown1990statistical}, takes the view that every sentence S in a source language has a possible translation T in the target language. Building on top of this fundamental assumption, SMT based approaches assign to each (S, T) sentence pair the probability $P(T|S)$, which is interpreted as the probability that sentence T is the translated equivalent in the target language of the sentence S in the source language. Accordingly, statistical approaches define the problem of Machine Translation as:

\begin{align}
    T &= \arg\max_{T} P(T|S) 
    \\
      &= \arg\max_{T} P(T)P(S|T)
\end{align}

The components $P(T)$ and $P(S|T)$ in the equation above are referred to as the Language Model of the target language, and the Translation Model respectively. Hereafter, we refer to the language model of the target language as the language model itself. Together, the language model and the translation model compute the joint probability of the sentences $S$ and $T$. The argmax operation over all sentences in the target language denotes the search problem and is referred to as the decoder. The decoder performs the actual translation - given a sentence $S$, it searches for a sentence $T$ in the target language with the highest probability $P(T|S)$. 

Since the current formulation requires a translation model for target language to the source language, an important question arises. Why can't the process to build this translation model be utilized to build the model that computes $P(T|S)$. This would eliminate the need for the language model of the target language, and can be used in conjunction with the decoder to get the translation of the original sentence. Brown \textit{et al.} \cite{brown1993mathematics} state this to be a means to get a well-formed sentence. To model $P(T|S)$ and use this for translation would require the probabilities to be concentrated over well-formed sentences in the target language domain. Rather, this is achieved through the joint usage of the language model and the translation model. Sentences which are not well-formed are expected to have a lower language model probability which offsets the necessity for the translation model to have their probabilities concentrated over well-formed sentences.

In the following sections, we first briefly review language models since they are modelled independent of the translation model and typically remain consistent across works in SMT. Thereafter, we review the research work in SMT categorized into two sections: Word-based SMT, and Phrase-based SMT.

\subsection{Language models}

Given a target string T of length m and consisting of words $t_1, t_2, \cdots, t_m$, we can write the language model probability $P(T)$ as:

\begin{align}
    P(T) = P(t_1, t_2, \cdots, t_m) = P(t_1) \prod_{i=2}^{m} P(t_i | t_{1:i-1})
\end{align}

This converts the language modelling problem into one that requires computing probabilities for a word given its history. However, computing these probabilities is infeasible since there could be too many histories for a word. Thus, this requirement is relaxed by truncating the dependence of the current word on a fixed subset of its history. In an n-gram model, it is assumed that the current word depends only on the previous (n-1) words. For example, in a trigram model, $P(w_i | w_{1:i-1}) = P(w_i | w_{i-1}, w_{i-2})$. These probabilities can now be computed through counting to get a Maximum Likelihood Estimate (MLE). There are other formulations of language models - ones that make use of neural networks, or formulate the problem as a maximum entropy language model. However, we won't delve deeper into language models since the majority of research in SMT is focused on different formulations of the translation model, but the reader can refer to the following resources for more information \cite{goodman2001bit} \cite{rosenfeld2000two} \cite{bengio2003neural}.

\subsection{Word-based SMT}

Post Warren Weaver's proposal in 1949 \cite{weaver1955translation} to use statistical techniques from the then nascent field of communication theory to the task of using computers to translate text from one language to another, research in the area lay dormant for a while. It wasn't until 1988 that Brown \textit{et al.} in \cite{brown1988statistical} outlined an approach to use statistical inference tools to solve the task. The authors argued that translation ought to be based on a complex glossary of correspondences of fixed locations. This glossary would map words as well as phrases (contiguous and non-contiguous) to corresponding translations. For example, the following could be the contents of a glossary mapping english words/phrases to their french counterpart: [word = mot], [not = ne pas], [seat belt = ceinture de sécurité]. The authors base their approach on the following decomposition of the task: 1) Partition the source text into a set of fixed locations, 2) Use the glossary and contextual information to select the corresponding set of fixed locations in the target language, and 3) Arrange the words of the target fixed location into a sequence that forms the target language. The proposed glossary in the paper is based on a model of the translation process $P(T|S)$, and comes to the critical conclusion that a probabilistic method is required to identify the corresponding words in the target and source sentence. To learn the parameters of this glossary, the paper introduces a concept of "generation pattern" which as we will see later is similar to the critical concept of alignment in machine translation. Since the authors were experimenting with English-French language pairs - languages with similar word order and therefore the translation being quite local - the fact that the proposed glossary did not incorporate this property motivated them to propose another formulation of the glossary - one that models the locality of the language pairs through distortion probabilities $P(k|h, m, n)$, where k refers to the $\text{k}^{\text{th}}$ word in T, h refers to the $\text{h}^{\text{th}}$ word in S, and m and n are the lengths of T and S. To the best of our knowledge, this work was the first to formalize the field of statistical machine translation, and though it provided only some intermediate results and not the translation examples it stimulated interest in the application of statistical methods to machine translation.

Two years later, in 1990, Brown \textit{et al.} in \cite{brown1990statistical} provided first experimental results for a statistical machine translation technique translating sentences in French to English. The proposed method translates 5\% of the sentences exactly to their actual translation, but if alternate and different translations are considered reasonable translations, the model's accuracy rises to 48\%. The authors further argue that this system reduces the manual work of translation by about 60\% as measured in units of key strokes. 

\begin{figure}[htp]
\begin{center}
  \includegraphics[scale=0.35]{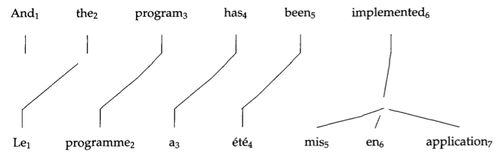}
  \caption{Alignment between two sentences.}
  \label{fig:alignment1}
\end{center}
\end{figure}

The translation model proposed by Brown \textit{et al.} in \cite{brown1990statistical} introduces the critical concept of \textit{alignment}. As defined by the authors, alignment between a pair of strings (S, T) indicates the origin in T of each word in S. One such alignment for the sentence pairs "Le programme a été mis en application" (S) and "And the program has been implemented" (T) is shown in figure \ref{fig:alignment1}. This particular alignment states that the origin of the word "the" in the english sentence lies in the word "Le", for "program" its "programme", and similarly "implemented" originates from the words "mis en application". Closely related to this concept of alignment is \textit{fertility}. Fertility is the number of words in S, that each word in T produces. Thus, for the same example, word "And" has fertility 0 (since it's not aligned with any word in french translation), "the" has fertility 1, and "implemented" has fertility 3.

Building on top of their previous work, Brown \textit{et al.} in \cite{brown1993mathematics} describe a set of five statistical models, each with a different model of the alignment probability distribution. Specifically, they modify their translation model to include the alignment variable A.

\begin{align}
    P(S|T) = \sum_A P(S, A|T)
\end{align}

For Models 1 and 2, the authors decompose $P(S, A|T)$ into three probability distributions: 1) distribution over the length of the target sentence, 2) alignment model defining a distribution over the alignment configuration, and 3) the translation probabilities over target sentence, given the alignment configuration and the source sentence. The main distinction between Models 1 and 2 is in their modelling of the alignment probabilities. Model 1 assumes a uniform distribution over all alignments for a sentence pair, while Model 2 uses a zero-order alignment model where alignments at different positions are independent of each other. Additionally, the trained parameters of Model 1 are used to initialize Model 2.

For Models 3, 4, and 5, the authors decompose $P(S, A|T)$ differently and parameterize fertilities directly. The generative process is broken into two parts: 1) Given T, compute the fertility of each word in T and a set of words in S which connect to it. This is called the tableau $\tau$, and 2) Permute the words in the tableau to form the source sentence S. This permutation is denoted as $\pi$. Accordingly, $P(S, A | T)$ is decomposed as:

\begin{align}
    P(S, A | T) = \sum_{\tau, \pi \in (S, A)} P(\tau, \pi | T)
\end{align}

$P(\tau, \pi | T)$ is further decomposed to result in the following parameters for Model 3 and 4: fertility probabilities, translation probabilities, and distortion probabilities. The main distinction between Model 3 and 4 lies in modelling of the distortion probabilities. Model 3 uses a zero-order distortion probabilities where the distortion for a particular position depends only on its current position and the lengths of T and S. Model 4 on the other hand, parameterizes these distortion probabilities by two sets of parameters: one to place the head of each word/phrase, and the other to place the rest of the words. This was done because Model 3 did not account well for the tendency of phrases to move around as a unit. 

Both of these models (Model 3 and 4) are however \textit{deficient}. The authors define a model to be deficient when it does not concentrate its probability over events of interest but rather distributes it over generalized strings. Model 5, which is the final of the proposed models, aims to avoid deficiency and does so by reformulating Model 4 by a suitably refined alignment model. Since each of the 5 proposed models have a particular decomposition of the translation model, the authors have tried to gain insights into the capabilities of these individual distributions as well as the final model itself. It is found that while the individual distributions model the particular events well, there is room for improvement in the model's capacity to translate.

The fundamental basis of the five models presented by Brown \textit{et al.} \cite{brown1993mathematics} was the introduction of a hidden alignment variable in the translation model. These alignment probabilities were then modelled differently in different models. Vogel \textit{et al.} \cite{vogel1996hmm} propose a new alignment model that's based on Hidden Markov Models (HMM) and aims to effectively model the strong localization effect when translating between certain languages (ex: for language pairs from Indoeuropean languages). The translation model $P(S|T)$ is accordingly broken down into two components: the HMM alignment probabilities and the translation probabilities. The key component of this approach is that it makes the alignment probabilities depend on the relative position of the word alignment rather than the absolute position. This HMM model is shown to result in smaller perplexities as compared to Model 2 by Brown \textit{et al.} \cite{brown1993mathematics} and also produces smoother alignments.

With the increased focus on research in alignment models, Och and Ney  \cite{och2000improved} present an annotation and evaluation scheme for word-alignment models. The proposed annotation scheme made it possible to explicitly annotate the ambiguous alignments along with the sure alignments. This provided an extra degree of freedom to the human annotators to generate reference alignments. To evaluate the performance of a word alignment model the authors propose an Alignment Error Rate which depends on the sure and ambiguous reference alignments, and the alignment produced by the model.

\subsection{Phrase-based SMT}
Despite the revolutionary nature of word-based systems, they still failed to deal with cases, gender, and homonymy. Every single word was translated in a single-true way, according to the machine. In phrase-based translation system there is no restriction of translating source sentence into target sentence word-by-word. This was a significant departure from word-based models - IBM models. In Phrase-based systems, a lexical unit is a sequence of words (of any length) as opposed to a single word in IBM models. Each pair of units (one each from source and target language) has a score or a `weight' associated with it. For example, a lexical entry could look like:\\

\centerline{(le chien, the dog,  0.002)}
\textit{
\begin{definition}
More formally, a phrase-based lexicon L is a set of lexical entries where each lexical entry is a tuple (f,e,g) where:
\begin{itemize}
    \item f is a sequence of one or more foreign language words
    \item e is a sequence of one or  more source language words
    \item g is a `score' of the lexical entry which is a real number.
\end{itemize}
\end{definition}
}

Phrase-based translation models improved the translation quality over IBM models and many researches tried to advance the state-of-the-art with these models. Och \textit{et al.}\cite{och2000improved} alignment template model can be reframed as a phrase translation system; Yamada and Knight\citep{yamada2001syntax} use phrase translation in a syntax based
translation system; Marcu \textit{et al.}\citep{marcujoint} introduced a joint-probability model for phrase translation. At its core, phrase-based translation system has a phrase translation probability table (defined above) to map phrases in source language to phrases in target language. The phrase translation table is learnt from word alignment models using bilingual corpus. We don't delve into the details of learning phrase lexicons from word alignments and encourage the reader to refer \citep{och2002statistical}, \citep{och2000improved} for details. We, instead, focus our discussions on modelling aspect of phrase-based systems and variations among different models.

Phrase-based systems decompose the translation probability defined in equation-2 as follows:

\begin{align}
    P(S|T) &= P(\bar{s}_1|\bar{t}_1)
    \\
    &= \prod_{i=1}^{I}\phi(\bar{s}_i|\bar{t}_i)d(a_i-b_{i-1})
\end{align}

$\bar{s}_1$ is the sequence of phrases in source sentence, $\bar{f}_1$ is the sequence of phrases in target sentence, $I$ is the number of sequences (in source sentence). $d(a_i-b_{i-1})$ is the relative distortion probability distribution, where $a_i$ denotes the start position of the source language
phrase that was translated into the \textit{i}th target language phrase, and $b_{i-1}$ denotes the end position of the source language phrase translated into the (\textit{i}-1)\text{th} target language phrase. $\phi(\bar{s}_i|\bar{t}_i)$ is the phrase translation probabilities (or equivalently phrase translation table) learnt from bilingual corpus and the distortion probability is either learnt or could be as simple as $\alpha^{|a_i-b_{i-1}-1|}$. The distortion probability distribution accounts for reordering of phrases in target language after they have been translated individually. Once all the factors (phrase translation tables, distortion distribution, language model) are learnt, the decoding operation (equation-1) generates translated sentences. The reader is encouraged to look at \citep{decode1}, \citep{decode2}, \citep{decode3} for more information on design of decoders and their nuances. 

Marcu \textit{et al.}\citep{marcujoint} present a different formulation of phrase-based model to learn the phrase transition table and distortion distribution. They argue that lexical correspondences can be established not only at the word level but also at the phrase level. They model the translation task as a joint probability model where the translations between phrases are learnt directly without using word alignment models. Their joint probability model is defined as:

\begin{align}
    p(E,F) = \sum\limits_{C \in \mathcal{C}|L(E,F,C)} \prod\limits_{c_i \in C}
    [t(\bar{e_i}, \bar{f_i}) * \prod\limits_{k=1}^{|\bar{f_i}|}
    d(pos(\bar{f}_{i}^{k}), pos_{cm}(\bar{e_i}))]
\end{align}

The generative story of this model is as follows:\\
\begin{enumerate}
    \item Generate of bag of concepts $C$ where each concept $c_i$ is the hidden variable.
    \item For each concept $c_i \in C$, generate a pair of phrases ($\bar{e_i}, \bar{f_i}$) according to the distribution $t(\bar{e_i}, \bar{f_i})$ where $\bar{e_i}$ and $\bar{f_i}$ each contain atleast one word.
    \item Order the phrases generated in each language so as to create two linear sequence of phrases; these sequences correspond to the sentence pair in bilingual corpus. This is modelled with $d(.)$ distribution.
\end{enumerate}

A set of concepts can be linearized into a sentence pair $(E, F)$ if $E$ and $F$ can be obtained by permuting the phrases  $\bar{e_i}$ and $\bar{f_i}$ that characterize all concepts $c_i \in C$.

To learn this model, they also propose a heuristics based learning algorithm. The model couldn't be learnt with the EM algorithm exhaustively as there are exponential number of alignments that can generate the sentence pair (E,F). They use French-English parallel corpus
of 100,000 sentence pairs from the Hansard corpus to train their model. Their model achieves boost in the BLEU score by 6 points compared to the IBM model 4 (with BLEU score of 22).

Och \textit{et al.}\citep{maxentropy} proposed a maximum entropy models for phrase-based translation where the translation probability is formulated as conditional log-linear model. The conditional probability of a sentence in target language given sentence in source language is:
\begin{align}
    Pr(\bar{e}_1^I|\bar{f}_1^J) &= p_{\lambda_{1}^{M}}(\bar{e}_1^I|\bar{f}_1^J)
    \\
    &=\frac{\text{exp}[\sum_{m=1}^{M}\lambda_{m}h_{m}(\bar{e}_1^I,\bar{f}_1^J)]}
    {\sum_{\bar{e'}_1^I}\text{exp}[\sum_{m=1}^{M}\lambda_{m}h_{m}(\bar{e'}_1^I,\bar{f'}_1^J)]}
\end{align}

In this framework, there is set of $M$ feature functions $h_{m}(\bar{e}_1^I,\bar{f}_1^J)$. For each feature function, there exists a model parameter $\lambda_m, m = 1,....,M$. The model is trained with the GIS (global iterative search) algorithm\citep{gis}. Since the normalization constant is intractable, it is approximated with highly-probable n sentences. The list of highly probable n sentences is computed by extended version from used search algorithm (Och \textit{et al.} \citep{och2000improved}) which approximately computes n-best list of translations. The main advantage of the maximum entropy model is that any feature function can be added easily (for eg., language model, distortion model, word penalty, phrase translation model) and the weights of these individual feature functions (models) can be learnt jointly.  They experiment with various feature functions including language model, word penalty, phrase translation dictionary and achieve state-of-the-art results on VERBMOBIL task which is a speech translation task in the domain of appointment scheduling, travel planning and hotel reservation.

\section{Neural Machine Translation}
In most statistical approaches to machine translation, the most crucial component of the system is the phrase transition model. It is either the joint probability of co-occurence of source and target language phrases, $P(e_i,f_i)$ or the conditional probability of generating a target language phrase given the source language phrase $P(e_i|f_i)$. Such models consider the phrases which are distinct on the surface as distinct units. Although these distinct phrases share many properties, linguistic or otherwise, they rarely share parameters of the model while predicting translations. There is no concrete notion of `phrase similarity' in such models. Besides ignoring phrase similarities, this leads to a very common problem of sparsity. It gets difficult for model to adapt itself to unseen phrases at test time. Finally, this makes it difficult to adapt such models to other similar domains. 

Continuous representations of linguistic units, be it character, word, sentence or document have shown promising results on various language processing tasks. One of early works which introduced this idea was proposed by Bengio \textit{et al.} \citep{bengio2003neural}. They model words with continuous fixed dimension word vectors using neural network and  achieve state-of-the-art results on language modelling task. It has also shown promising results in dealing with sparsity issue. Collobert \textit{et al.} \citep{collobert} have shown that continuous representations for words are able to capture the syntactic, semantic and morphological properties of the words. Continuous representations for characters have also shown notable results in language modelling task as proposed by Sutskever \textit{et al.} \citep{sutskeverrnn}. Recently, continuous representations have been proposed for phrases and sentences and have been shown to carry task-dependent information to help downstream language processing tasks (Grefenstette \textit{et al.} \citep{grefenstette}, Socher \textit{et al.} \citep{socher10}, Hermann \textit{et al.} \citep{herman13}).

The approaches discussed above make use of neural networks to model continuous representations of linguistic units. Deep neural networks have shown tremendous progress in computer vision (eg., Krizhevsky \textit{et al.} \citep{krizhevsky}) and speech recognition (eg., Hinton \textit{et al.} \citep{hintonspeech} and Dahl \textit{et al.} \citep{dahlspeech}) tasks. Since then, they have also been successfully applied to solve many NLP tasks like paraphrase detection (Socher \textit{et al.} \citep{socherparaphrase}) and word embedding extraction (Mikolov \textit{et al.} \citep{mikolov}). Neural networks have also been applied to advance the state-of-the-art in statistical machine translation. Schwenk \citep{Schwenk2012ContinuousST} summarizes usage of feedforward neural networks in the framework of phrase-based SMT system.

\subsection{Preliminary}
In the next two sections (4.1.1 and 4.1.2), we discuss some background work which is common to almost all the neural machine translation systems.

\subsubsection{Recurrent Language Model}

A recurrent neural network (RNN) is a neural network that consists of a hidden state h and an optional output y which operates on a variable length sequence x = ($x_1$... $x_T$). At each time step t, the hidden state $h_t$ of the RNN is updated by: 
\begin{align}
    h_t = f(h_{t-1}, x_t)
\end{align}
where the f is non-linear activation function which is usually implemented with LSTM cell (Hochreiter and Schmidhuber \citep{lstm}). Using softmax function with vocabulary size V, an RNN can be trained to predict the distribution over $x_t$ given the history of words ($x_{t-1}, x_{t-2},...x_1$) at each time step t. By combining
the probabilities at each time step, we can compute the probability
of the sequence x (eg: target language sentence) using
\begin{align}
    p(x) = \prod_{t=1}^{T}p(x_t|x_{t-1},...,x_1)
\end{align}
which is called the Recurrent Language Model (RLM).

\subsubsection{Encoder-Decoder Architecture}

Though RNN Encoder-Decoder architecture was proposed by Cho \textit{et al.} \citep{choemnlp14} for a machine translation task, it remains the base model for most of the NLP sequence-to-sequence models (and especially machine translation). We discuss this model in its general form here, and delve into details of different neural architectures in next section.

\begin{figure}[htp]
\begin{center}
  \includegraphics[scale=0.6]{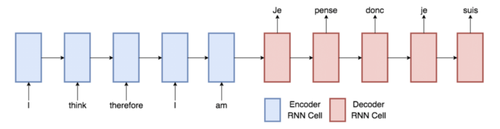}
  \caption{An encoder-decoder architecture}
  \label{fig:encoder-decoder}
\end{center}
\end{figure}

An encoder-decoder neural model (figure \ref{fig:encoder-decoder}), from a probabilistic perspective, is a general method to learn the conditional distribution over variable length sequence given yet another variable length sequence, e.g. $p(y_1,...,y_{T^{'}}|x_1,...,x_{T})$. An encoder is an RNN which reads each symbol in input sequence (x) one word at a time till it encounters end-of-sequence symbol. The hidden state of the RNN at the last time step is the summary c of the whole input sequence. The decoder  operates very similar to RLM discussed  previously except that the hidden state of the decoder $h_t$ now depends on the summary c too. Hence the hidden state of the decoder at time step t is calculated by 
\begin{align}
    h_{t} = f(h_{t-1}, y_{t-1}, c)
\end{align}
and the conditional distribution of the next symbol (for e.g. next word in target language sentence given source language sentence) is
\begin{align}
    p(y_t|y_1,...,y_{t-1}) = g(h_t, y_{t-1}, \textbf{c})
\end{align}
 The two components of the encoder-decoder model are jointly trained to maximize the conditional log-likelihood 
 \begin{align}
 \frac{1}{N}\sum_{n=1}^{N}\text{log } p_{\theta}(y_n|x_n)
 \end{align}
 
We now describe some of the Neural Machine Translation (NMT) methods proposed recently.

\subsection{NMT Methods}

Motivated from success of deep neural networks and their ability to represent a linguistic unit with a continuous representation, Kalchbrenner \textit{et al.} \citep{cnnencoder} propose a class of probabilistic translation models, Recurrent Continuous Translation Model (RCTM) for machine translation. The RCTM model has a generation aspect and a conditional aspect. The generation of a sentence in target language is modelled with target Recurrent language model. The conditioning on the source sentence is modelled with a Convolutional Neural Network (CNN). In their model, CNN takes a sentence as input and generates a fixed size representation of this source sentence. This representation of source sentence is presented to the Recurrent Language Model to produce the translation in target language. The entire model (CNN and RNN) is trained jointly with back-propagation.

To the best of our knowledge, this is the first work which explores the idea of modelling the task of machine translation entirely with neural networks, with no component from statistical machine translation systems. They propose two CNN architectures to map source sentence into fixed size continuous representation. Though CNN architectures have shown tremendous success in image space, these architectures were first explored extensively in text space in this paper. We, therefore, discuss these architectures in detail here.

\begin{figure}[htp]
\begin{center}
  \includegraphics[scale=0.5]{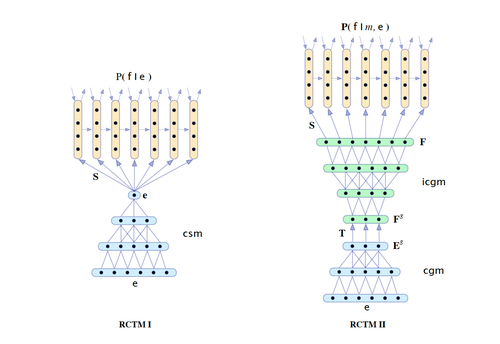}
  \caption{A graphical depiction of the two RCTMs. Arrows represent full matrix transformations while lines are
vector transformations corresponding to columns of weight matrices.}
  \label{fig:rctm}
\end{center}
\end{figure}

The Convolutional Sentence Model (CSM)
creates a representation for a sentence that is progressively built up from representations of the n-grams in the sentence. The CSM architecture embodies a hierarchical structure, similar to parse trees, to create a sentence representation. The lower layers in the CNN architecture operate locally on n-grams and the upper layers act increasingly globally on the entire sentence. The lack of need of parse tree makes it easy to apply these models to languages for which parsers are not available. Also, generation of the sentence in target language is not dependent on one particular parse tree. Similar to CSM, the authors propose another CNN model called Convolutional n-gram model (CGM). The CGM is obtained by truncating the CSM at the level where n-grams are represented for the chosen
value of n. The CGM can also be inverted (icgm) to obtain a representation
for a sentence from the representation of its n-grams.  The transformation icgm unfolds the n-gram representation onto a representation of a target sentence with m target words (where m is also predicted by the network according to the Poisson distribution). The pictorial representation of two models is shown in figure \ref{fig:rctm}. 

The experimentation is performed on a bilingual corpus of 144953 pairs of sentences less than 80 words in length from the news commentary section of the Eighth Workshop on Machine Translation (WMT) 2013 training data. The source language is English and the target language is French. A low perplexity value achieved by RCTMs on test set as compared to IBM models (model 1-4) suggests that continuous representations and the transformations between them make up well for the lack of explicit alignments. To make sure that RCTM architecture (with CGM) doesn't just take bag-of-words approach, they change the ordering of the words in the source sentence and train their model. This model achieves much lower perplexity values which proves that the model is indeed sensitive to source sentence structure. They also compare the performance of the RCTM model with cdec system. cdec employs 12 engineered features including,
among others, 5 translation models, 2 language model features and a word penalty feature (WP). RCTM models achieve comparable performance (marginally better) than the cded system on BLEU score. The results indicate that the RCTMs are able to learn both translation and language modelling distributions without explicitly modelling them.

Cho \textit{et al.} \citep{choemnlp14} propose a RNN encoder-decoder architecture very similar to the one proposed above but with one major difference. While Kalchbrenner \textit{et al.} \citep{cnnencoder} use CNN to map a source sentence into a fixed-sized continuous representation, Cho \textit{et al.} \citep{choemnlp14} use an encoder RNN to map source sequence into a vector. However, they use this architecture to learn phrase translation probabilities. The training is done on phrase translation pairs extracted in the phrase-based translation system. The model re-scores all the phrase-pairs probabilities which are used as additional features in log-linear phrase based translation system. They use WMT’14 translation task to build English/French SMT system coupled with features from encoder-decoder model. With quantitative analysis of the system (on BLEU score), they show that baseline SMT system's performance was improved when RLM was used. Additionally, adding features from proposed Encoder-Decoder architecture increased the performance further suggesting that signals from multiple neural systems indeed add up and are not redundant. They later perform qualitative analysis of their proposed model to investigate the quality of the target phrases generated by model. The target phrases (given a source phrase) proposed by model look more visually appealing than the top target phrases from translation table. They also plot the phrase representations (after dimensionality reduction) on 2-d plane and show that the syntactically and semantically similar phrases are clustered together.

While Cho \textit{et al.} \citep{choemnlp14} proposed an end-to-end RNN architecture, they use it only to get additional phrase translation table to be eventually used in the SMT based system. Sutskever \textit{et al.} \citep{seqtoseq} gave a more formal introduction to the sequence-to-sequence RNN encoder-decoder architecture. Though their motivation was to investigate the ability of very-deep neural networks at solving seq-to-seq problem, they run their experiments on machine translation task to achieve their goals. They proposed the architecture very similar to Cho \textit{et al.} \citep{choemnlp14} with three major architecture changes: 1) they used LSTM cells in encoder and decoder RNN, 2) they trained their system on complete sentence pairs and not just phrases, 3) they used stacked LSTMs (with 4-6 layers) in both decoder and encoder. Finally, they also reverse the source sentences in the training data and train their system on reversed source sentences (keeping the target language sentences in their original order). They don't provide a clear motivation of why they did so, but informally, reversing the source sentence helps in capturing local dependencies around the word from either direction. Their experimentation results (on WMT-14 English/French MT dataset) show that reversing the source sentence achieves higher BLEU score on test set than the model where no reversing was done. Though their model doesn't beat the state-of-the-art MT system, it achieves performance very close to the latter. Their model doesn't employ any attention methods or bi-directional RNN (which is used by the state-of-the-art system). This suggests that deep models indeed help in seq-to-seq learning with RNN encoder-decoder architecture.

Neural machine translation has shown very promising results for many language pairs. Despite that, it has only been applied to only formal texts like WMT shared task. Luong \textit{et al.} \citep{luongstanford} study the effectiveness of NMT systems in spoken language domains by using IWSLT 2015 dataset. They explore two scenarios: NMT \textit{adaptation} and NMT for \textit{low resource translation}. For NMT adaptation task, they take an existing state-of-the-art English-German system\citep{luongattention}, which consists of 8 individual models trained on
WMT data with mostly formal texts (4.5M sentence pairs. They further train on the English-German spoken language data provided by IWSLT 2015 (200K sentence pairs). They show that NMT adaptation is
very effective: models trained on a large amount of data in one domain can be finetuned on a small amount of data in another domain. This boosts the performance of an English-German NMT system by 3.8 BLEU points. For NMT \textit{low resource translation} task, they use the provided English-Vietnamese parallel data (133K sentence pairs). At such a small scale of data, they could not train deep LSTMs with 4 layers as in the English-German case. Instead, they opt for 2-layer LSTM models with 500-dimensional embeddings and LSTM cells. Though their system is little behind the IWSLT baseline (baseline's BLEU score is 27.0 and their model's BLEU score is 26.4), it still shows that NMT systems are quite effective in other domains too, and not just formal texts.

\subsubsection{Attention Mechanisms}
A potential issue with this encoder–decoder approach is that a neural network needs to be able to compress all the necessary information of a source sentence into a fixed-length vector. This may make it difficult for the neural network to cope with long sentences, especially those that are longer than the sentences in the training corpus. Cho \textit{et al.} \citep{cho14b} showed that indeed the performance of a basic encoder–decoder deteriorates rapidly as the length of an input sentence increases. Bahdanau \textit{et al.} \citep{bengioattention} proposed an attention mechanism to deal with this issue. They propose a model where the source sentence is not encoded into one fixed length vector. Instead, the encoder maps the source sentence into sequence of vectors and decoder chooses a subset of these vectors at each time step to generate tokens in the target language. We now discuss this model more formally. In the proposed architecture, the conditional probability is defined as:
\begin{align}
    p(y_{i}|y_1,...y_{i-1},x) = g(y_{i-1}, s_i, c_i)
\end{align}
where $s_i$ is an RNN hidden state for time i, computed by:
\begin{align}
    s_i = f(s_{i-1}, y_{i-1}, c_i)
\end{align}
The context vector $c_i$ depends on a sequence of annotations ($h_1,... h_{T_{x}}$) to which an encoder maps the input sentence. The context vector $c_i$ is, then, computed as a weighted sum of these annotations $h_i$:
\begin{align}
    c_i=\sum_{j=1}^{T_{x}}\alpha_{ij}h_j
\end{align}
The weight $\alpha_{ij}$ of each annotation is computed by 
\begin{align}
    \alpha_{ij} = \frac{\text{exp}(e_{ij})}{\sum_{k=1}^{T_{x}}\text{exp}(e_{ik})}
\end{align}
where $e_{ij} = a(s_{i-1},h_j)$ is an alignment model, implemented with feed-forward neural network. The alignment scores how well the inputs around position \textit{j} and output at position \textit{i} match. We can understand the approach of taking a weighted sum of all the annotations as computing an expected annotation, where the expectation is over possible alignments. They use bidirectional RNN for encoder and a unidirectional RNN for decoder, and use the bilingual, parallel corpora provided by ACL WMT ’14 task (English/French). With their experiments on source sentences of different lengths, they show that the performance of the conventional encoder-decoder drops quickly when the sentence length increases beyond 30. On the other hand, the proposed model remains less volatile to the sentence length and continues to achieve good performance on long sentences too. They also plot alignment visualizations for each target word produced by decoder. The visualizations show that English and French languages are highly monotonic, which indeed is the case with these languages.

Luong \textit{et al.} \citep{luongattention} propose two attention approaches: a global approach which always attends to all source words and a local one that only looks at a subset of source words at a time. The global approach is very similar to the one proposed by Bahdanau \textit{et al.} \citep{bengioattention} but is architecturally simpler than the latter. The local attention can be viewed as a blend between soft and hard alignment approaches proposed by Xu \textit{et al.} \citep{xuattentions}. The local attention model is computationally less expensive and differentiable, making it easier to implement and train. We don't discuss the global attention model here as it is very similar to the one proposed by Bahdanau \textit{et al.} \citep{bengioattention}, and direct the reader to the original paper for details and comparisons. We focus our discussions on local attention model which is the major contribution of this work. In local attention, the model first generates the aligned position $p_t$ for each target word at time t. The context vector $c_t$ is then derived as a weighted average over the set of source hidden states within the window [$p_{t-D}, p_{t+D}$];D is empirically selected. Unlike the global approach, the local alignment vector at is now fixed-dimensional, i.e., $\in \mathbb{R}^{2D+1}$. The model predicts the aligned position $p_t$ as follows:
\begin{align}
    p_t = S.\text{sigmoid}(v_p^\text{T}\text{ tanh}(W_ph_t))
\end{align}
$W_p$ and $v_p$ are the model parameters which will
be learned to predict positions. $S$ is the source sentence length. To favor alignment points near pt,
Gaussian distribution is used and centered around $p_t$. The alignment weights are now defined as:
\begin{align}
    a_{t}(s) = \text{align}(h_t, \bar{h}_s) \text{ exp} \left(-\frac{(s-p_t)^2}{2\sigma^2}\right)
\end{align}
The align function can be as simple as a dot product or can be learned with feed-forward neural net. The standard deviation is empirically set as $\sigma=\frac{D}{2}$ and $s$ is an integer within the window centered around $p_t$. They evaluate the effectiveness of the model on the WMT translation tasks between English and German in both directions, and use WMT’14 training data, and newstest2014 (2737 sentences) and newstest2015 (2169 sentences) as their test data. Apart from achieving higher BLEU scores (even on longer sentences) than the baseline system NMT system (without attention) and other conventional SMT approaches, they also visualize the quality of the alignments produced by the model during decoding. After learning, they extract only one-to-one alignments by selecting the source word with the highest alignment weight per target word and compare it with gold alignment data provided by RWTH for 508 English-German Europarl sentences. They use the alignment error rate (AER) to measure the alignment quality of the model. The results show that they were able to achieve AER scores comparable to the one-to-many alignments obtained by the Berkeley aligner (Liang \textit{et al.} \citep{aligner}).

\subsubsection{Addressing OOV} 
A significant weakness in conventional NMT systems is their inability to correctly translate very rare words: end-to-end NMTs tend to have relatively small vocabularies with a single symbol that represents \textit{unk} every possible out-of-vocabulary (OOV) word. Standard phrase-based systems, on the other hand, do not suffer from this problem to same extent as NMTs as they make use of explicit alignments and phrase tables which allows them to memorize the translations of extremely rare words. 

Jean \textit{et al.} \citep{bengiorare} propose a method based on importance sampling that allows them to use large target language vocabulary without increasing training complexity. They divide the training set into multiple individual sets, each having its own target vocabulary V'. More concretely, before training begins, each target sentence is sequentially examined and  unique words are accumulated till number of unique words reach predefined threshold $\tau$. The
accumulated vocabulary will be used for this partition of the corpus during training. The process is repeated until the end of the training set is reached. An intrigued reader can refer the original paper for more formal description of the model and the training procedure. The proposed approach is evaluated on English to French and English to German task. Bilingual parallel corpora from WMT'14 is used for training the model. Apart from showing the efficiency of the proposed model through comparable BLEU scores with the state-of-the-art WMT'14 submitted model, they propose heuristic-based changes to the traditional NMT decoder to make it sample efficiently from extremely large target vocabulary. 

Yet another method was proposed recently to address the OOV problem. Luong \textit{et al.} \citep{luangrareword} train an NMT system on data that is augmented by the output of a word alignment algorithm, allowing the NMT system to emit, for each OOV word in the target sentence, the position of its corresponding word in the source sentence. This information is later utilized in a post-processing step that translates every OOV word using a dictionary. Experiments on the WMT’14 English to French translation task show that this method provides improvement of up to 2.8 BLEU points over an equivalent NMT system that does not use this technique. 

\section{Current Research}
Kaiser \textit{et al.} \citep{onemodel} recently proposed an interesting neural network architecture - `One Model to learn them all'. Its a Multi-Model architecture that can simultaneously learn many tasks across domains. At its core, it has four components - modality nets, encoder, IO mixer and decoder. Modality nets (one each for all types of data - text, speech, audio, image) map input into a representation. Encoder takes this representation and processes it with attention blocks and mixture-of-experts\citep{moe}. Decoder, in a similar fashion, produces output representation which is given to the respective modality net to produce the output. Both encoder and decoder are built with convolutional blocks. Their experiments on various tasks (including machine translation) show that the model performs, if not at par yet, but close to the state-of-the-art systems on individual tasks. They also show that attention and mixture-of-experts blocks, designed for textual data (especially machine translation) doesn't hurt the performance of other completed unrelated tasks like classification on ImageNet\citep{imagenet}.

Another area of current research is related to the requirement of a large parallel corpus to train NMT systems. The lack of such corpus for low-resource languages (e.g. Basque) as well as for combinations of major languages (e.g. German-Russian) poses a challenge for such systems. Artetxe \textit{et al.} \citep{artetxe2017unsupervised} propose an unsupervised approach to neural machine translation which relies solely on a monolingual corpus. The system architecture is a standard encoder-decoder setup with the encoder shared across the two decoders along with attention. The encoder contains pre-trained cross-lingual word embeddings which are kept fixed during training. Ideally, this architecture can be trained to encode a given sentence using the shared encoder and decode it using the appropriate decoder, but it is prone to learn trivial copying task. To circumvent this, the authors propose using denoising and on-the-fly backtranslation. Denoising randomizes the order of the words to force the network to learn meaningful information about the language, and the on-the-fly backtranslation translates text from the available monolingual corpus to the other language to get a pseudo-parallel sentence pair. This architecture improves over the baseline scores by at least 40\% on both German-English  and French-English translation. This unsupervised approach is also shown to improve with the availability of small parallel corpus.

\section{Conclusion}
Machine translation has been an active area of research within the field of AI for many years. Statistical machine translation, with the advent of IBM models (model 1-4) paved way for advanced approaches based on phrase-based and syntax-based models. These methods have shown tremendous progress in many language pairs and have been successfully deployed in large scale systems, like Google translate (up until 2014). Over the past couple of years, Neural Machine Translation has taken the front seat in this task. Owing to their ease of learning, their ability to model complex feature functions and their striking performance in translating major languages of the world, NMT systems have become natural choice for researchers to study their behavior, the feature space they learn, and the effect of variations in architectures. Despite that, much needs to be done both from modelling perspective and architecture changes. We believe that a unified architecture similar to the one proposed in `One model to learn them all' holds potential in benefiting from multiple tasks learnt simultaneously. Finally, we are also seeing unsupervised methods being applied to learn MT systems from just one language. This research area is especially important considering the number of languages in the world and the limited amount of labelled data available for them.

\newpage

\bibliographystyle{plainnat}
\bibliography{test}

\end{document}